\documentclass{article}

\usepackage{PRIMEarxiv}
\usepackage{multirow}
\usepackage[utf8]{inputenc} 
\usepackage[T1]{fontenc}    
\usepackage{hyperref}       
\usepackage{url}            
\usepackage{booktabs}       
\usepackage{amsfonts}       
\usepackage{nicefrac}       
\usepackage{microtype}      
\usepackage{lipsum}
\usepackage{fancyhdr}       
\usepackage{graphicx}       
\usepackage{amsmath}
\graphicspath{{media/}}     

\title{Adapting Tree-Structured Speculative Decoding to DeepSeek-V4 for Efficient Inference}

\author{
  Changxu Liu$^{1,2}$, Zhaogeng Li$^1$\\
  $^1$Baige AI Team, Baidu Inc. \ $^2$Fudan University \\
}

\begin{document}
\maketitle

\begin{abstract}
Repeated execution of the target model during autoregressive decoding is a major source of LLM inference latency. Unlike linear speculation, which follows a single candidate chain, tree-structured speculation retains multiple branches from shared prefixes; under the same budget, this broader coverage can improve acceptance and efficiency. Adapting it to DeepSeek-V4 is nontrivial: its CSA/HCA online compressed attention concentrates the difficulty on the target-verify side, where branches diverging from a shared prefix compress into different states, breaking cross-branch state consistency. We integrate tree-structured speculative decoding into the DeepSeek-V4-Flash pipeline via branch-aware causal verification, temporary state isolation, and accepted-path state refresh, keeping verification and compressed-state updates consistent across branches. Across budgets D=5 to D=8, batch sizes 1 to 64, and three datasets (GSM8K, MBPP, ShareGPT), tree speculation achieves a higher accepted length than the matched linear configurations in all settings (e.g., at D=8 about 2.83--3.41 versus 2.39--2.84) and improves throughput in nearly all configurations---marginal only at the smallest budget---by up to about 18.5\%. More importantly, the gains follow stable, transferable regularities: the relative gain grows with the budget and is most pronounced for less predictable workloads at small-to-medium batch sizes, while beyond a certain budget throughput plateaus and decouples from the still-rising accepted length. These results show that retaining multiple candidate paths under the same budget can effectively improve DeepSeek-V4 decoding efficiency, and offer experience for adapting speculative decoding to future models with compressed, sparse, or structured context representations.
\end{abstract}


\section{Introduction}
Autoregressive decoding remains a major source of latency in large language model inference, since the target model must repeatedly execute a decoding step for each newly generated token. This issue becomes more pronounced for long-context models, where attention computation and context management introduce additional costs. DeepSeek-V4 addresses the long-context efficiency problem with a new compressed-attention design that combines Compressed Sparse Attention (CSA) and Heavily Compressed Attention (HCA) \cite{xu2026deepseek}. By compressing the context history online across sequence positions, this design reduces both the amount of context that must be stored and the amount of context involved in subsequent attention computation. Nevertheless, token generation in DeepSeek-V4 remains autoregressive, leaving repeated target-model execution as an important source of inference latency.

Speculative decoding provides a way to reduce this cost \cite{leviathan2023fast}. In EAGLE-3/MTP-style autoregressive speculative decoding, a draft model generates a group of future candidates, and the target model verifies them in a subsequent forward pass \cite{li2026eagle3,gloeckle2024better}. More recently, DeepSeek's DSpark~\cite{cheng2026dspark} improves the draft side---obtaining longer accepted lengths at lower draft-decode cost and adapting the verification depth to the serving load---a direction orthogonal to, and composable with, the verify-side focus of this work. The number of tokens accepted in each verification cycle determines how many target-model forward passes are amortized across the generated tokens, and thus directly affects decode throughput. A linear speculative strategy follows a single candidate chain. When an early draft prediction is incorrect, the remaining suffix of the chain may become unusable. Tree-structured speculation eiqrtuwy    instead preserves multiple alternatives from shared prefixes. For a comparable verification budget, this broader candidate coverage can lead to a higher acceptance rate and create an opportunity to reduce the effective decoding latency.

This motivates our study of adapting tree-structured speculative decoding to DeepSeek-V4. The adaptation is challenging because DeepSeek-V4 does not represent its decoding history as independent per-token key-value entries: its CSA and HCA paths build compressed representations online and reuse them in later attention. When speculative candidates form multiple branches, the logical ancestry of tokens no longer matches their physical order, and the compressed history must stay consistent with the candidate structure throughout drafting, verification, and acceptance. The central challenge is therefore not constructing a candidate tree, but preserving tree-speculation semantics on the target-verify side, within an attention system that actively compresses and reorganizes its context history.

Based on this background, this work focuses on the following aspects:
\begin{itemize}
    \item We investigate tree-structured speculative decoding as a practical approach to reducing DeepSeek-V4 inference latency within the EAGLE-3/MTP-style autoregressive speculative decoding framework.
    \item We analyze the compatibility challenges introduced by the interaction between tree-structured speculative execution and DeepSeek-V4's online, sequence-level compressed attention, with particular attention to maintaining consistent context history during drafting, verification, and acceptance.
    \item We adapt the existing DeepSeek-V4 inference stack to support tree-based speculative verification and acceptance, and optimize the execution path to control the additional coordination and state-management overhead introduced by dynamic tree processing.
    \item We evaluate the adapted implementation against matched linear speculative-decoding configurations in terms of accepted length and decode throughput. Based on these results, we distill empirical regularities---across verification budgets, batch sizes, and datasets---into engineering insights for future models and inference strategies with structured context representations.
\end{itemize}

Our evaluation shows that tree-structured speculation achieves a higher accepted length than the matched linear configurations across all tested settings. Although tree execution adds coordination and state-management overhead, the extra accepted tokens per verification cycle can outweigh this cost under appropriate configurations, improving decode throughput. We further distill empirical regularities across budgets, batch sizes, and datasets---most notably that the relative gain grows with the verification budget and is most pronounced for less predictable workloads at small-to-medium batch sizes. Beyond the specific DeepSeek-V4 implementation, this study offers practical experience for adapting speculative decoding to future models with compressed, hierarchical, sparse, or otherwise structured context representations.

\section{Background} \label{sec:headings}

\subsection{Linear and Tree-Structured Speculative Decoding}
The speculative decoding strategy considered in this work follows the EAGLE/MTP-style autoregressive formulation. A lightweight drafting component predicts multiple future tokens, which are subsequently verified by the target model in a single forward pass. By accepting multiple draft tokens in one verification cycle, speculative decoding amortizes the cost of repeatedly executing the target model  \cite{leviathan2023fast}. EAGLE-style methods perform autoregressive drafting with model features, while multi-token prediction provides a related mechanism for predicting several future tokens  \cite{li2026eagle3,gloeckle2024better,li2024eagle}.

Within this framework, linear and tree-structured speculation organize draft candidates in different ways. Linear speculation produces a single candidate chain, in which each draft token depends on its predecessor. During verification, an error near the beginning of the chain can invalidate the remaining suffix and limit the number of accepted tokens.

Tree-structured speculation retains multiple continuations from shared prefixes. Candidate tokens are organized according to their parent-child relationships, allowing alternative sequences to be verified together while preserving the causal dependency between each token and its ancestors. An incorrect prediction on one branch does not necessarily invalidate candidates on other branches. This token-tree organization and parallel verification paradigm has been explored in prior systems such as SpecInfer \cite{miao2024specinfer}, while EAGLE-2 studies the construction of dynamic draft trees based on candidate confidence \cite{li2024eagle2}. Compared with a single linear chain, retaining multiple alternatives can increase the probability of finding a longer accepted path and improve the number of accepted tokens per verification cycle.

\begin{figure}[htb]
    \centering
    \includegraphics[width=\linewidth]{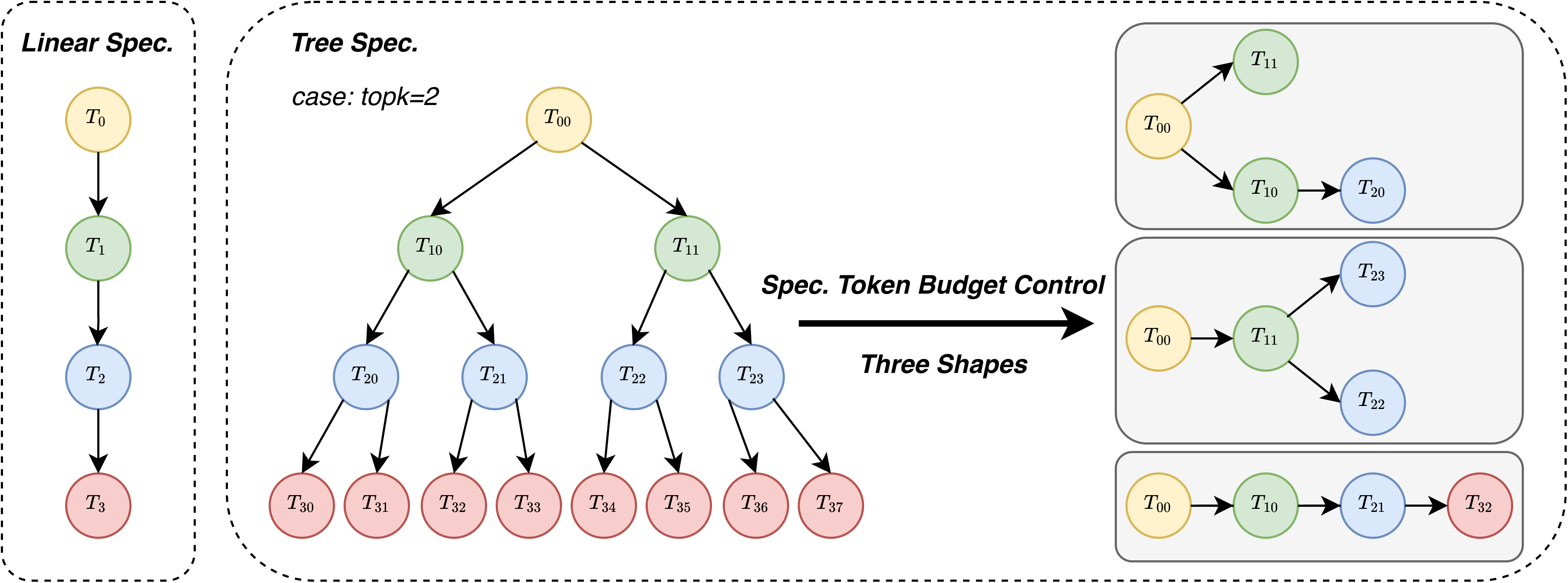}
    \caption{Linear and tree-structured EAGLE/MTP-style speculative decoding with budget-aware pruning. The example uses $\mathrm{top}\text{-}k=2$ and three drafting steps; other parameter settings may produce different tree shapes.}
    \label{tree_spec_background}
\end{figure}

The size of the candidate tree is controlled by a verification-token budget. After the draft tree is expanded, pruning is applied to retain promising candidate nodes while keeping the number of tokens verified by the target model within the budget. This budget-aware selection balances candidate coverage against verification cost. Sequoia studies a related problem of optimizing the tree structure under speculation and hardware constraints \cite{chen2024sequoia}. An excessively large verification set may increase the arithmetic workload and shift target-model verification from a memory-bound regime toward a compute-bound regime, thereby reducing the latency benefit of further tree expansion.

Fig.~\ref{tree_spec_background} illustrates an example with top-k = 2 and three drafting steps. Under this configuration, pruning the expanded candidate tree can produce the three representative shapes shown on the right. These shapes are specific to the illustrated configuration; different top-k values, drafting depths, or verification budgets may produce other tree structures.

Tree-structured speculation introduces a trade-off between candidate coverage and execution cost. Although retaining multiple branches may improve acceptance, it also incurs additional overhead for candidate organization, tree-aware verification, and accepted-path management. In this report, tree speculation is considered an established extension of EAGLE/MTP-style speculative decoding, and the focus is on adapting this strategy to the compressed-attention architecture of DeepSeek-V4.

\subsection{DeepSeek-V4 Inference Architecture}
DeepSeek-V4 introduces an attention architecture built around Compressed Sparse Attention (CSA) and Heavily Compressed Attention (HCA) \cite{xu2026deepseek}. Unlike conventional attention mechanisms that maintain an independent key-value representation for every historical token, CSA and HCA construct compressed representations of the context online across sequence positions. These representations are subsequently used by later attention computation, reducing both context storage and the amount of historical context involved in attention.

This design differs from token-wise latent KV compression, which primarily reduces the representation dimension of cached keys and values while retaining an approximately one-to-one correspondence between sequence positions and cache entries. In DeepSeek-V4, compression also operates along the sequence dimension: information from multiple token positions contributes to a shorter and structured representation of the context history. As decoding proceeds, this compressed history evolves together with the generated sequence.

The broader idea of maintaining compressed context during online generation has been explored in prior work. For example, Compressed Context Memory recursively compresses accumulated key-value representations into a compact memory for online language-model interaction \cite{kim2024compressed}. DeepSeek-V4, however, integrates sequence-level compression directly into its inference-time attention architecture through CSA and HCA. These mechanisms provide context representations at different compression granularities, forming a multi-resolution history that helps control the storage and computation cost of long-context inference.

From an inference-system perspective, this architecture creates a distinction between the logical token history and the physical representation consumed by attention. The logical history remains an autoregressive token sequence, whereas the physical history includes compressed context states derived from multiple sequence positions. The attention state is therefore not merely a collection of independent per-token cache entries; its validity also depends on the sequence prefix from which the compressed representations were constructed.

This property is particularly relevant to speculative decoding. Linear speculation extends a single tentative sequence, whereas tree speculation produces multiple branches that share a prefix but represent mutually exclusive continuations. Applying tree-structured speculation to DeepSeek-V4 therefore requires the candidate topology, causal visibility, and evolving compressed context history to remain semantically consistent. The specific correctness and execution challenges arising from this interaction are discussed in the next section.

\section{Challenges}
Extending EAGLE/MTP-style speculative decoding from a linear sequence to a candidate tree changes more than the organization of draft tokens. Multiple branches must be verified under different causal relationships, DeepSeek-V4's internal states must stay consistent with the eventually accepted sequence, and the extra work of dynamic tree execution must stay light enough for improved acceptance to translate into end-to-end gains. These requirements lead to three main challenges.

\subsection{Branch-Aware Causal Verification and Isolation}
In linear speculative decoding, draft tokens form a single sequence whose dependencies follow a conventional causal order. Tree-structured speculation breaks this assumption: each candidate should attend to the accepted context and the ancestors along its own branch, but not to candidates on unrelated branches. Since tokens from different branches are verified together, their physical order in the verification batch no longer reflects their logical dependencies, so the target model must preserve the tree topology during causal verification. Similar requirements arise in prior token-tree verification methods such as SpecInfer and in the tree-attention verification used by Medusa \cite{cai2024medusa}.

Multiple branches also require isolation. Before verification completes, every branch is tentative: intermediate results from one branch must not affect other branches or the already accepted state. After verification, the result is no longer a single accepted-prefix length as in a linear chain; the system must identify a specific path through the tree, retain its results, and discard those of rejected branches. The challenge is thus to support tree-aware causal visibility and branch isolation within an execution path originally designed for linear sequences.

\subsection{Compressed-State Consistency and Accepted-Path Refresh}
Isolation is harder in DeepSeek-V4 because speculative tokens affect more than conventional per-token KV entries. CSA and HCA build compressed context representations online as the sequence grows, so branches that share an accepted prefix may still produce different compression results after they diverge. Until the accepted path is known, these results cannot be treated as permanent context, or information from a rejected branch would leak into the compressed history and affect later decoding. Token-level verification alone is therefore insufficient: the compressed state must also match the accepted sequence.

Once verification completes, the state must be refreshed along the accepted path—preserving accepted candidates and invalidating rejected ones—so that the token history, attention cache, and the compression states of CSA and HCA all represent the same logical prefix. The main difficulty is preventing speculative updates from prematurely modifying persistent context while still producing a complete, consistent state after acceptance.

\subsection{Execution Overhead and Optimized-Inference Compatibility}
Tree-structured speculation adds work beyond a linear chain: candidate-tree construction, branch-aware verification metadata, temporary-state management, accepted-path extraction, and state refresh all add latency to a decoding cycle. Dynamic token-tree systems such as DySpec likewise identify irregular tree construction as a measurable overhead and optimize it accordingly \cite{xiong2024DySpec}. A higher acceptance rate helps only when the extra accepted tokens outweigh these costs.

Dynamic tree execution can also conflict with optimized inference paths: varying tree shapes produce different token counts, attention patterns, and state-update requirements, whereas mechanisms such as CUDA Graph replay and preallocated memory rely on stable shapes and predictable layouts, and extra data movement or synchronization on the critical path further erodes the advantage. The verification-token budget must likewise stay bounded so that processing extra candidates does not incur disproportionate computation, though detailed hardware-aware budget tuning is outside our scope. The main performance challenge is to preserve tree speculation's acceptance advantage while keeping branch management, state handling, and dynamic execution from dominating the decoding cycle.

Together, these challenges show that adapting tree-structured speculation to DeepSeek-V4 is not simply a matter of applying a tree-shaped attention mask: the system must jointly preserve branch-aware causal semantics, maintain compressed-state consistency along the accepted path, and control the cost of dynamic tree execution. The next section describes how these requirements are addressed.

\section{Our Approach}
This work adapts existing tree-structured speculative decoding to DeepSeek-V4 by organizing speculative inference into three stages: draft extend, draft decode, and target verify. The main adaptation is required in target verify, where multiple candidate branches must be verified while maintaining correct causal visibility and consistent CSA/HCA states.

The accepted prefix is treated as the persistent reference state, whereas speculative branches remain temporary until the accepted path is selected. The adapted execution flow is finally integrated into SGLang’s existing speculative inference pipeline.

\subsection{Three-Stage Speculative Forward}
A speculative forward extends the accepted prefix, computes candidate representations, and verifies the resulting candidate tree. These operations correspond to draft extend, draft decode, and target verify.

\textbf{Draft extend expands the accepted prefix into candidate branches and organizes them as a tree.} The final tree shape depends on the draft outputs, branch-related parameters, and the verification-token budget. Thus, this stage organizes the candidate topology rather than independently determining a fixed tree shape.

\textbf{Draft decode computes draft-side representations under the tree topology.} Each candidate token follows the causal path defined by its ancestors. DeepSeek-V4 already provides the required tree-attention support in its FlashMLA/attention path, so the adaptation mainly supplies the candidate layout and corresponding attention metadata.

\textbf{Target verify validates the candidate branches and selects the accepted path.} Before verification, a verify preprocess converts the candidate tree into a verification plan. This plan describes how to extract the candidate token chains and guides both attention and compression processing.

Chains sharing a prefix may reuse the prefix state, but branch-specific chains must be compressed independently after branching. Combining different chains in one compression operation would mix mutually exclusive sequence histories and produce semantically inconsistent compressed tokens. The verification plan therefore defines both the candidate-chain organization and the corresponding compression boundaries. After verification, the system selects the accepted path and commits only the states associated with that path. The updated accepted prefix then becomes the starting point for the next speculative forward.

\subsection{Temporary Branch States and Compressed-State Handling}

The state-management design distinguishes token-level KV updates from the path-dependent states generated by online sequence compression. This distinction makes CSA/HCA more difficult to support than SWA.

\textbf{SWA mainly requires token-level KV state refresh.} After verification, KV entries belonging to the accepted path are retained, while entries from rejected branches are discarded or invalidated. Since SWA does not perform the same type of online sequence compression, it does not require chain-specific compressed tokens or compression-intermediate buffers.

\textbf{CSA/HCA require isolated compression states for each candidate chain.} A candidate chain may produce its own compressed tokens, compressed-attention states, and intermediate compression results. These states remain temporary during verification, and only the state corresponding to the accepted path is used to update the persistent context.

A scratch pad stores these temporary branch states until target verification is complete. This prevents speculative branches from being written directly into the persistent cache before their validity is known. The scratch pad is also necessary because DeepSeek-V4 uses a page-based KV cache pool that can be viewed as one logical layer and three physical layers with shared page-level pointers. Directly writing multiple speculative branches into this persistent pool would complicate page allocation, state ownership, and branch reclamation. Temporary storage avoids these operations during branch exploration. Once the accepted path is selected, the system refreshes its token history, KV states, CSA/HCA compressed states, and relevant compression-intermediate buffers. Rejected branch states are not carried into the next decoding iteration.

\subsection{Performance Optimization}
The main optimizations reduce unnecessary cache operations, adapt refresh timing to different compression paths, and move metadata processing toward the device.

\textbf{Scratch-pad execution reduces persistent cache traffic.} Speculative branches do not repeatedly allocate and update persistent cache pages before acceptance. If there is only one candidate chain, or if the accepted chain is already the latest cached chain, the corresponding refresh can be skipped.

\textbf{C4 and C128 use different refresh schedules.} C4 is compressed more frequently and therefore requires timely updates to its compressed state and intermediate buffers. C128 is compressed less frequently, so some cache writes can be delayed and combined after the accepted path is known.

\textbf{Device-side metadata processing improves execution overlap.} Candidate indices, chain boundaries, and verification-plan metadata can be prepared or transformed on the device where appropriate. This reduces host-side control overhead and allows host scheduling, GPU metadata processing, and target-model execution to overlap.

\textbf{CUDA Graph coverage is bounded by the candidate-tree configuration.} The theoretical graph limit is estimated from the number of candidate chains and the relevant draft and verification parameters. The limit must cover common tree shapes without causing excessive memory use from unused graph capacity.

\textbf{Draft-side candidate layout and metadata are reused by target verification whenever possible.} The overall goal is to ensure that the acceptance-length benefit of tree speculation translates into lower end-to-end latency rather than being offset by cache, metadata, synchronization, or memory-management overhead.

\subsection{Integration into SGLang}
\begin{figure}[htb]
    \centering
    \includegraphics[width=\linewidth]{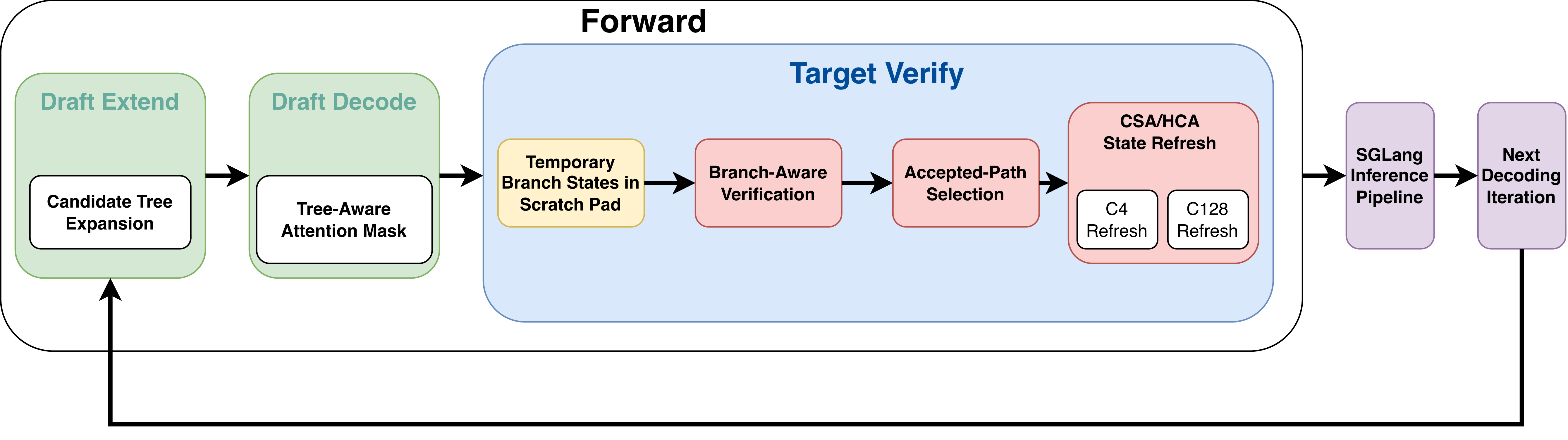}
    \caption{Three-Stage Speculative Forward with Accepted-Path State Management. The adapted speculative forward integrated into SGLang, from candidate-tree construction and verify preprocessing to branch-aware target verification, accepted-path state refresh, and the next decoding iteration.}
    \label{sglang_tree_spec_pipeline}
\end{figure}

The complete execution flow is integrated into SGLang as one speculative forward with three consecutive stages, as illustrated in Fig.~\ref{sglang_tree_spec_pipeline}. SGLang first provides the current accepted prefix and persistent cache state to the speculative module. The module then performs draft extend, which expands the accepted prefix into candidate branches, followed by draft decode, which computes candidate representations under the tree topology.

Before target-model execution, the candidate tree enters verify preprocess. At this point, the system constructs the verification metadata and extracts the candidate chains required by target verification. The resulting plan is passed to target verify, where the target model performs branch-aware causal verification. Temporary results from different candidate chains are kept in the scratch pad, preventing speculative states from being written directly into the persistent cache. After verification, the system performs accepted-path selection. Only the states associated with the selected path are retained. The corresponding token states, KV states, and compressed-attention states are then refreshed, including the C4/C128-specific state updates shown in the figure. Rejected branches are discarded and do not affect the persistent context.

The refreshed accepted prefix and cache state are returned to the SGLang inference pipeline, which starts the next decoding iteration from the committed state. In this way, the figure summarizes how candidate generation, verification preprocessing, target verification, temporary state management, accepted-path commitment, and subsequent scheduling form one complete execution loop.

\section{Experiments}
\subsection{Setup and Metrics}
We evaluate the DeepSeek-V4-Flash model on an eight-GPU NVIDIA machine, comparing linear and tree-structured speculation across a range of verification budgets and batch sizes. We use verification budgets D=5, 6, 7, 8 and batch sizes 1, 2, 4, 8, 16, 32, 64.

The evaluation covers three tasks with distinct generation characteristics: GSM8K (math word problems, with relatively regular and predictable reasoning chains), MBPP (Python programming, i.e. code generation), and ShareGPT (open-domain multi-turn dialogue, with divergent content and low predictability). These three datasets are chosen to observe how tree-structured speculation behaves across workloads ranging from highly predictable to highly divergent.

Speculation configurations are named in the order "draft steps / top-k / verification budget." For example, s4\_k1\_d5 denotes 4 draft steps, top-k=1, and a verification budget of 5 (top-k=1 degenerates to a single linear chain), whereas s4\_k2\_d5 denotes the same number of steps and budget but with top-k=2, i.e. a tree configuration. Under a matched verification budget, the linear and tree configurations have the target model verify the same number of candidate tokens; therefore any difference between them stems solely from how the candidates are organized (single chain vs. tree), forming a controlled comparison.

We report two metrics:
\begin{itemize}
    \item \textbf{Accepted length:} the average number of tokens accepted per speculative decoding round, measuring candidate quality. Reported both as absolute values (Tab.~\ref{tab:accept_len_abs}) and as the relative change of tree over the matched-budget linear configuration (Fig.~\ref{accept_len}).
    \item Decode throughput: the amount of effective decoding completed per unit time. Reported as the relative change of tree over the matched-budget linear configuration (Fig.~\ref{throughput}).
\end{itemize}
Since absolute quantities such as accepted length vary with hardware platform, model, and draft quality, the analysis below focuses primarily on the trends this dataset exhibits stably across budgets, configurations, batch sizes, and datasets.






\subsection{Accepted Length}\label{sec_accept_len}
\begin{table}[htbp]
\centering
\caption{Accepted length (absolute, averaged over batch sizes 1--64) of linear
and tree-structured speculation under matched verification budgets $D$. Within
each budget, the linear ($k{=}1$) and tree ($k{=}2$) configurations verify the
same number of candidate tokens. Accepted length is nearly invariant to batch
size (spread $<0.04$ across all tested batch sizes).}
\label{tab:accept_len_abs}
\begin{tabular}{clccc}
\toprule
Budget & Config. & GSM8K & MBPP & ShareGPT \\
\midrule
\multirow{3}{*}{$D{=}5$}
 & s4\_k1\_d5 (linear) & 2.836 & 2.663 & 2.389 \\
 & s3\_k2\_d5          & 3.135 & 2.943 & 2.681 \\
 & s4\_k2\_d5          & 3.154 & 2.935 & 2.667 \\
\midrule
\multirow{4}{*}{$D{=}6$}
 & s5\_k1\_d6 (linear) & 2.841 & 2.669 & 2.392 \\
 & s3\_k2\_d6          & 3.219 & 3.038 & 2.757 \\
 & s4\_k2\_d6          & 3.274 & 3.044 & 2.754 \\
 & s5\_k2\_d6          & 3.256 & 3.030 & 2.741 \\
\midrule
\multirow{4}{*}{$D{=}7$}
 & s6\_k1\_d7 (linear) & 2.845 & 2.672 & 2.392 \\
 & s4\_k2\_d7          & 3.361 & 3.129 & 2.809 \\
 & s5\_k2\_d7          & 3.348 & 3.108 & 2.798 \\
 & s6\_k2\_d7          & 3.329 & 3.098 & 2.795 \\
\midrule
\multirow{4}{*}{$D{=}8$}
 & s7\_k1\_d8 (linear) & 2.842 & 2.678 & 2.393 \\
 & s5\_k2\_d8          & 3.406 & 3.163 & 2.839 \\
 & s6\_k2\_d8          & 3.394 & 3.153 & 2.834 \\
 & s7\_k2\_d8          & 3.387 & 3.153 & 2.830 \\
\bottomrule
\end{tabular}
\end{table}

\begin{figure}[htb]
    \centering
    \includegraphics[width=\linewidth]{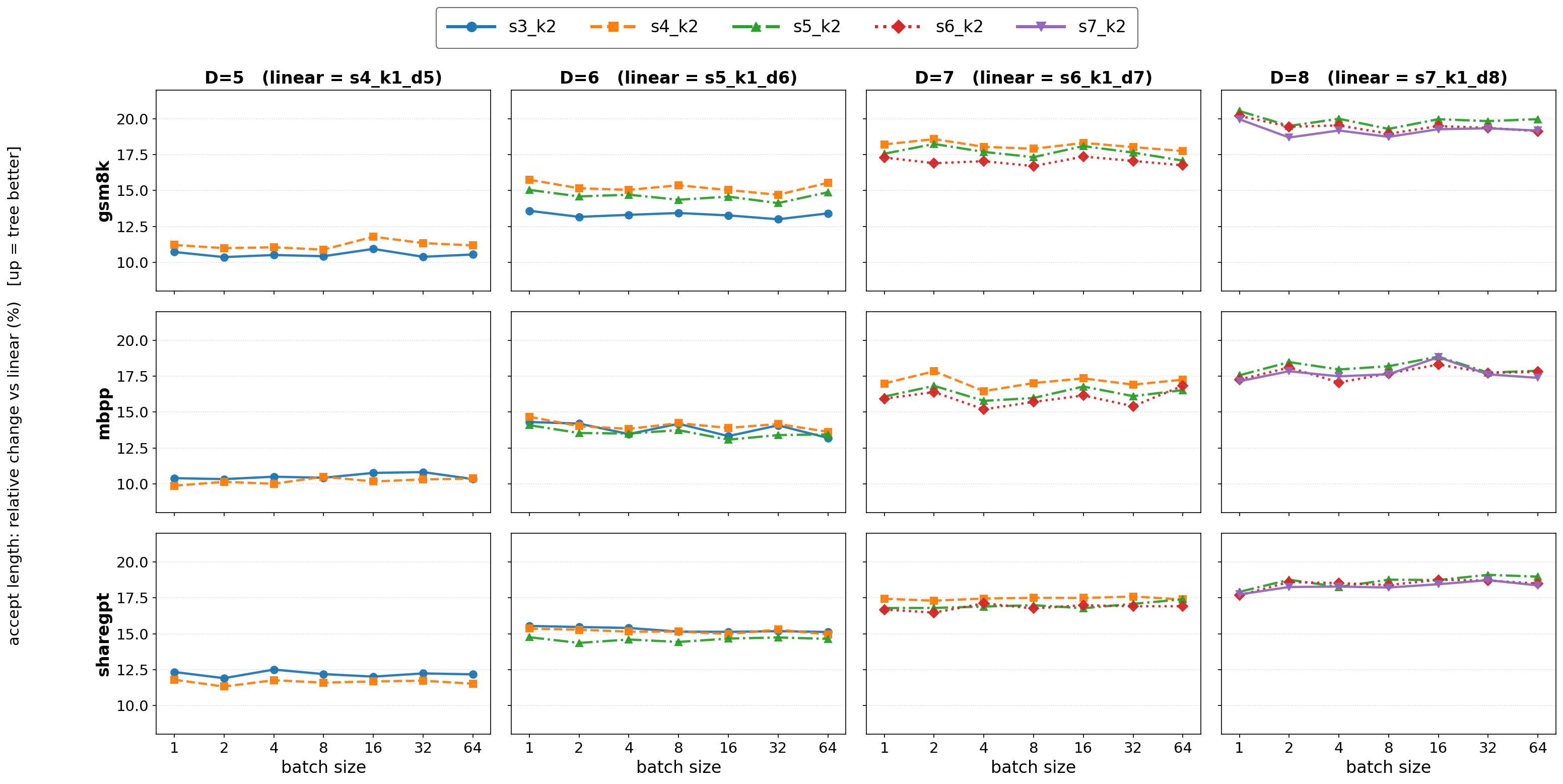}
    \caption{Relative change (\%) in accepted length of tree speculation over the matched-budget linear baseline (topk$=1$ at the same budget $D$); the baseline is the $0\%$ line and higher is better. Rows $=$ datasets, columns $=$ budget $D$; within each subplot, curves $=$ different draft-step configurations and the $x$-axis $=$ batch size.}
    \label{accept_len}
\end{figure}

Fig.~\ref{accept_len} shows the relative change in accepted length for tree vs. linear speculation (the y-axis is the percentage improvement of tree over linear; up means tree is better), and Tab.~\ref{tab:accept_len_abs} gives the corresponding absolute accepted-length values. As Fig.~\ref{accept_len} and Tab.~\ref{tab:accept_len_abs} show, across all budgets and batch sizes the tree configuration achieves a higher accepted length than its matched linear counterpart, and several stable patterns emerge.
\begin{enumerate}
    \item The relative gain grows monotonically with the budget D. The improvement of tree over linear rises from about +11.0\% at D=5 to +14.4\% at D=6, +17.1\% at D=7, and +18.6\% at D=8. Part of the budget always goes to maintaining chain depth; only the surplus beyond that can spread candidate width over a shared prefix—and width is where tree wins, keeping multiple branches at positions the draft is unsure of. Smaller budgets leave little surplus (the tree nearly degenerates to linear); larger budgets give width room to pay off. The linear baseline barely moves across budgets (Tab.~\ref{tab:accept_len_abs}: about 2.84 / 2.67 / 2.39 at every D), so this rising gain is entirely tree-side—and its per-budget increment shrinks (+3.4\% / +2.6\% / +1.6\%) as both chains near the draft's reliable-prediction horizon.
    \item Accepted length is nearly invariant to batch size. For a fixed configuration and dataset, it fluctuates by less than 0.04 from bs=1 to bs=64, since it depends only on draft quality and candidate structure, not on batching. This cleanly decouples the analysis in Sec.~\ref{sec_throughput}: throughput's bs dependence is an execution-level effect, not an acceptance effect. 
    \item  Deepening the chain buys no extra acceptances; the tree offers another way to raise the ceiling. For a fixed budget, "shallow-and-wide" vs. "deep-and-narrow" differ within about 2\% (e.g. s3\_k2, s4\_k2, s5\_k2 at D=6 are nearly identical): with today's draft models, a longer chain grows less reliable the further it reaches. Since depth is exhausted, the tree instead keeps more branches within the same budget to push the acceptance ceiling higher—which is exactly where it keeps adding value while drafts remain too weak for reliable long-range prediction.
    \item The quality improvement holds robustly across tasks. Absolute accepted length follows
    a stable GSM8K > MBPP > ShareGPT ordering (both linear and tree), tracking task
    predictability. Yet the relative gain lands in a similar range across all three
    (usually within one or two points at the same budget), so the tree's benefit does not
    depend strongly on task type. MBPP consistently shows the lowest relative gain, while GSM8K and ShareGPT are comparable (GSM8K edging slightly ahead at larger budgets). The
    differences are small, so we do not read a strong baseline-dependent trend into them.
\end{enumerate}

\subsection{Decode Throughput}\label{sec_throughput}
\begin{figure}[htbp]
    \centering
    \includegraphics[width=\linewidth]{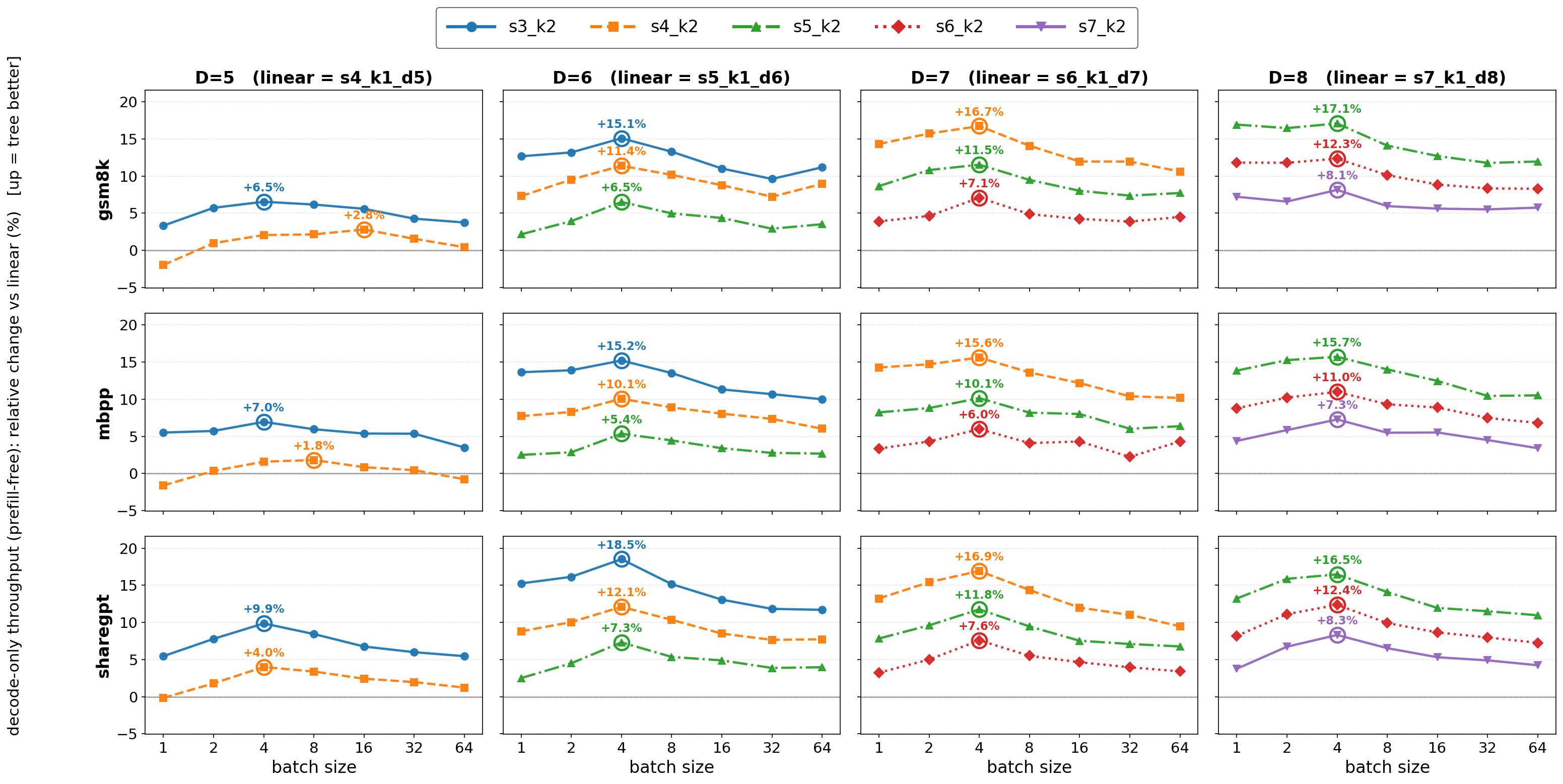}
    \caption{Relative change (\%) in decode throughput of tree speculation over the matched-budget linear baseline (topk$=1$ at the same budget $D$); the baseline is the $0\%$ line and higher is better. Rows $=$ datasets, columns $=$ budget $D$; within each subplot, curves $=$ different draft-step configurations and the $x$-axis $=$ batch size.}
    \label{throughput}
\end{figure}

Fig.~\ref{throughput} shows the relative change in decode throughput of tree over the matched-budget linear configuration (up means tree is better). As Fig.~\ref{throughput} shows, tree-structured speculation yields positive throughput gains across most tested configurations, with only a few small-budget (D=5) cases near break-even. The magnitude of the gain varies clearly and interpretably across budgets, configurations, batch sizes, and datasets.

\begin{enumerate}
    \item Across budgets: gains grow with D and are marginal when D is small. Average gain rises from about +3–5\% at D=5 to +8–9\% at D=6, then plateaus at roughly +9–10\% through D=7 and D=8 (essentially on par); at D=5 a few configurations (e.g. s4\_k2\_d5 at bs=1) are near break-even or slightly negative. Throughput gain is essentially the accepted-length improvement minus the tree's fixed overhead (candidate organization, branch-aware verification metadata, temporary state, accept-path refresh). At small budgets the acceptance gain is too small to offset that overhead; it only emerges once the budget is large enough. This is telling against accepted length, which keeps climbing through D=8 (Sec.~\ref{sec_accept_len}): beyond D$\approx$6 the extra acceptance the tree gains per added unit of budget shrinks while its overhead grows, so the two cancel—accepted length and throughput decouple, and more budget buys quality but not speed.
    \item Within a budget: moderately fewer draft steps tend to buy higher throughput. At D=6, the average throughput gain drops from about +13\% (s3\_k2) to +9\% (s4\_k2) to +4\% (s5\_k2), with D=7 the same. Since accepted length is nearly step-insensitive (Sec.~\ref{sec_accept_len}), the difference falls on overhead: fewer steps means a shallower, wider tree and fewer serial draft forward passes. This is not "fewer is always better," though—too shallow a tree lacks candidate depth and accepted length suffers, and the optimal depth shifts deeper as the draft grows stronger. So "fewer steps wins" holds only within our range, where width's marginal benefit currently exceeds depth's.
    \item Across batch sizes: an inverted-U, peaking at small-to-medium bs. Gains peak around bs=4 and taper toward bs=64 (at D=6, roughly bs1 +8\% → bs4 +11\% → bs64 +7\%). Since accepted length is bs-independent (Sec.~\ref{sec_accept_len}), this is purely execution-level: at small bs, decode is memory-bound, so accepting more tokens per round directly cuts target-model forward passes; at large bs, the target forward is compute-bound and already amortized across sequences, so the extra verified candidates instead contend for compute. Our general-purpose multi-GPU setup is not tuned for high-concurrency serving, which adds some fluctuation at large bs.
    \item The hardest-to-predict workloads see the largest speedup. All three datasets gain, with ShareGPT the highest and GSM8K/MBPP slightly lower; the global peak of +18.5\% occurs here (s3\_k2\_d6, bs=4). This mirrors the accepted-length trend (Sec.~\ref{sec_accept_len}): where a single chain most often fails early, the tree's width has the most waste to recover.
\end{enumerate}
Taking these dimensions together, whether the tree's accepted-length advantage can be converted into throughput depends on the match among budget scale, the allocation between depth and width, and the batch size and workload characteristics at deployment: the budget must be large enough to leave room for spreading width, the depth must be commensurate with draft capability, and small-to-medium batch sizes and less-predictable workloads are more likely to see pronounced gains. Beyond D$\approx$6–7, extra budget brings diminishing throughput returns despite continued accepted-length gains, so pushing the budget higher offers little on this setup. When these conditions are well matched, our adaptation on DeepSeek-V4-Flash improves decode throughput by up to about 18.5\% over linear speculation.

\section{Discussion and Future Work}
\paragraph{Tree Speculation and Draft-Side Schemes (e.g., DSpark) Are Orthogonal and Complementary.}
DSpark~\cite{cheng2026dspark} is a representative recent draft-side scheme: its strength lies on the draft (proposal) side, obtaining a longer accepted length at a smaller draft-decode cost and adaptively adjusting the verification depth at runtime according to the online environment. This is orthogonal to our work, which improves the verify side: replacing a single chain with a tree so that, under the same budget, more candidate branches are verified. Draft-side efficiency (DSpark) and verify-side width (tree structure) are two independent dimensions that do not conflict: as long as a scheme like DSpark can propose (or be extended to propose) a tree, our tree-verification machinery applies. The two can advance separately and their gains compose.

\paragraph{The Difficulty and Overhead Concentrate on the Target-Verify Side and Grow with the Complexity of the Attention Paradigm.}
The real adaptation difficulty of tree speculation lies on the target-verify side, and its cost depends on whether the attention state diverges across branches once they split from a shared prefix. Dense attention keeps the history explicitly per token, so a tree only needs a causal mask over the shared KV, introducing no extra cross-branch divergence and keeping the cost contained. Linear attention must maintain and roll back a recurrent state, but this cost is inherent to speculative decoding and borne by linear speculation as well, so the tree adds little. What truly makes the difference is compressed attention (e.g., CSA): once branches diverge they compress into different states, a cross-branch difference unique to tree speculation that requires branch-aware causal verification, temporary state isolation, and accepted-path state refresh. DeepSeek-V4's CSA/HCA is one such instance, and the compressed-attention paradigm persists into DeepSeek-V4.1-Flash~\cite{deepseekai2026deepseekv41flash}, which upgrades CSA to CSA2 and drops HCA yet still follows this compression path. Hence a transferable observation: the net gain of a tree over the matched linear configuration is roughly (the extra acceptance from a wider tree) $-$ (the incremental verification cost from cross-branch state divergence), with the latter largest under the compressed paradigm---explaining why gains are marginal at small budgets and saturate at large ones.

\paragraph{Adaptive, Workload-Aware Tree Shaping.}
At a fixed budget, accepted length is nearly insensitive to the number of draft steps, so shallower and wider trees tend to achieve higher throughput; yet this advantage has a lower bound, and the optimal depth is coupled to draft capability---the stronger the draft, the deeper the optimal depth shifts. More importantly, draft-side depth adaptivity (as in DSpark) and verify-side overhead are coupled through the shape of the tree: the proposal side determines the tree's shape, and the verification cost in turn depends on it, so depth and width should be decided jointly by both sides. In addition, the current draft tree is pruned mainly by accumulated scores; beyond scores, future work could incorporate branch depth, path structure, and inter-candidate dependencies, and---combining workload predictability with the runtime batch size (our data show that less predictable tasks yield larger gains and that gains follow an inverted-U over batch size)---adaptively decide whether to enable tree speculation and how wide the tree should be.

\paragraph{Applicability Conditions and Outlook for Tree Speculation.}
Whether tree speculation pays off depends on two conditions holding simultaneously. First, the draft's acceptance capability has become the bottleneck---our data confirm this precisely: merely deepening the chain buys no additional acceptances, indicating that the bottleneck lies in the draft rather than the budget, and here widening offers another path to raise the acceptance ceiling. Second, the target side can verify the tree cheaply enough. As more models adopt compressed, sparse, and structured context representations, the latter condition---the target-verify side---is becoming the decisive factor. Advancing target-verify adaptation, as this work does, is therefore what keeps tree speculation valuable in the face of ever-stronger draft schemes.

\section{Conclusion}
This work adapts tree-structured speculative decoding to DeepSeek-V4 and integrates it into the DeepSeek-V4-Flash pipeline, using branch-aware causal verification, temporary state isolation, and accepted-path state refresh to resolve the state-consistency challenges of CSA/HCA online compression. Across verification budgets from D=5 to D=8, tree speculation achieves a higher accepted length than linear speculation and improves decode throughput in nearly all configurations (marginal only at the smallest budget), by up to about 18.5\%. Beyond the absolute numbers, these gains follow stable, transferable patterns---growing with the budget and depending on the depth--width allocation and the workload. Since its overhead concentrates on the target-verify side and grows with cross-branch state divergence (largest under compressed attention such as CSA), tree speculation is orthogonal and complementary to draft-side schemes such as DSpark, and pays off most when the draft is the bottleneck. Future work will co-design the tree's depth, width, and activation with draft capability and runtime workload, and continue advancing target-verify adaptation as compressed, sparse, and structured contexts become mainstream, so as to improve scalability in long-context serving.

\bibliographystyle{unsrt}  
\bibliography{references}

\end{document}